\documentclass[10pt,twocolumn,letterpaper]{article}

\usepackage[pagenumbers]{cvpr} 
\usepackage{multirow}
\usepackage{algorithm}
\usepackage{algpseudocode}
\usepackage{subcaption}
\definecolor{cvprblue}{rgb}{0.21,0.49,0.74}
\usepackage[pagebackref,breaklinks,colorlinks,allcolors=cvprblue]{hyperref}
\usepackage{cuted}
\usepackage{amsmath}

\def\paperID{21} 
\def\confName{CVPR}
\def\confYear{2026}

\title{PRS-Med: Position Reasoning Segmentation in Medical Imaging}

\author{Quoc-Huy Trinh$^{1,2}$, Minh-Van Nguyen$^{3}$, Jun Zeng$^{4}$, Debesh Jha$^{5,\dagger}$, Ulas Bagci$^{2,\dagger}$\\
$^{1}$Aalto University \quad $^{2}$Northwestern University \quad $^{3}$Technical University of Denmark\\
$^{4}$Chongqing University of Posts and Telecommunications \quad $^{5}$University of South Dakota\\
$^{\dagger}$Co-advising
\quad \url{https://huyquoctrinh.github.io/prsmed/}
}

\begin{document}
\maketitle
\begin{abstract}
Prompt-based medical image segmentation has rapidly emerged, yet existing methods rely on explicit prompts like bounding boxes and struggle to reason about the spatial relationships essential for clinical diagnosis. While general-domain models attempt complex coordinate regression, these approaches often lack the structured reliability required for medical applications. In this work, we introduce PRS-Med, a unified framework that adopts an elegant, clinical-first approach to position reasoning segmentation. By utilizing a medical vision-language model integrated with a segmentation decoder, PRS-Med mimics the structured "search patterns" used by radiologists to identify pathologies within specific anatomical zones. To support this robust reasoning, we present the Medical Position Reasoning Segmentation (PosMed) dataset, comprising 116,000 expert-validated, spatially grounded question-answer pairs across six imaging modalities. Unlike previous brittle attempts at spatial reasoning, PosMed leverages a scalable, deterministic pipeline validated by board-certified radiologists to ensure clinical accuracy. Extensive experiments demonstrate that our zone-based reasoning not only improves segmentation accuracy (mean Dice improvements up to +31.2\%) but also provides a high-confidence interpretability layer that outperforms state-of-the-art complex reasoning models. By prioritizing functional reliability over unnecessary technical complexity, PRS-Med offers a practical and scalable baseline for the next generation of intelligent medical assistants.
\end{abstract}    
\section{Introduction}
\label{sec:intro}
In the medical field in general and oncology in particular, doctors typically make diagnoses by examining potential tumor locations and types to evaluate tissue conditions. This makes position reasoning and segmentation visualization crucial for supporting early and accurate diagnoses. As medical assistant agents become more common, models like LLaVA-Med \cite{llavamed}, Med-MoE \cite{medmoe}, HuatuoGPT \cite{chen2024huatuogpt}, and MedVLM-R1 \cite{medvlm-r1} have been developed and shown potential in the diagnostic. However, they still face a challenge in position identification when doctors often need to identify unknown tumor locations to make a decision for their diagnosis. Additionally, the position information can help doctors recognize the growing tumors, which leads to more effective diagnosis and treatment. This technology can also help clinics create automated screening systems, reducing manual costs.

In the natural image domain, several works such as LISA~\cite{lisa}, LLM-SEG~\cite{llm-seg}, and SegLLM~\cite{wang2024segllm} have addressed the challenge of reasoning for segmentation, achieving notable success in enhancing object reasoning, identifying object positions through segmentation, and providing simple reasoning about objects. However, these Multimodal-LLMs are not well-trained on medical imaging, making their application to this field difficult. This is due to the complex nature of medical content and the difficulty of boundary learning in medical segmentation, which out-of-domain models struggle with. For position reasoning segmentation, a VLM's vision model needs to be well-trained on medical images to effectively distinguish and localize tumors and anatomies for reasoning. 

We present \textbf{PRS-Med}, a framework for \textbf{P}osition \textbf{R}easoning \textbf{S}egmentation in medical imaging. This is a unified method that uses a Multimodal-LLM to perform position-reasoning segmentation from simple questions or commands. Our model outputs both a textual description and a segmentation mask that highlights the tumor location. PRS-Med acts as an intelligent assistant, answering a doctor's questions and visually indicating the position of tumors or anatomical structures in an interpretable way. In addition, general-domain models strive for pixel-perfect coordinate regression; PRS-Med prioritizes Clinical Action Zones, ensuring that the spatial reasoning mimics the structured language of board-certified radiologists. Our contributions are as follows:

\begin{itemize}

    \item To address the lack of datasets for position reasoning in medical imaging, we create and release the Medical Position Reasoning Segmentation (PosMed) dataset pipeline. This pipeline can build a coarse position reasoning dataset, designed to generate diverse, spatially grounded question-answer pairs in the medical context. 

    \item We present PRS-Med, a position reasoning segmentation framework for Medical Imaging. It performs spatially-aware tumor segmentation using implicit natural language prompts.

    \item We are open-sourcing the dataset pipeline, model, and codebase to help the community develop spatially-aware multimodal LLMs in medical imaging.

\end{itemize}
\section{Related Work}
\label{sec:related_work}
\textbf{Medical Image Segmentation.} Traditional Medical image segmentation has long relied on fully supervised CNN-based architectures like U-Net~\cite{ronneberger2015u} and its variants, such as ResUNet++~\cite{jha2019resunet++}, nnu-net~\cite{isensee2018nnu}, DoubleUNet~\cite{jha2020doubleu}, TransResUNet~\cite{tomar2022transresu}, Swin-UNet~\cite{cao2022swin}, and MEGANet~\cite{bui2024meganet}. More recently, several promptable segmentations have emerged as a response to the growing demand for interactive and context-aware medical AI. MedSAM~\cite{MedSAM}, SAM-Med2D~\cite{cheng2023sammed2d} adapts the Segment Anything Model (SAM)~\cite{kirillov2023segment} to medical settings, supporting box- and point-prompted segmentation. However, MedSAM and SAM-Med2D still lack semantic understanding of positional cues within free-form text. In contrast, BiomedParse~\cite{biomedparse} directly uses text prompts to infer object shapes and positions, learning implicit position priors. Despite the promising results of previous works, they \textbf{lack the ability to segment with contextual reasoning}, nor does it support implicit or conversational prompts beyond class names prompt input. In this work, we propose the PRS-Med, which allows the MLLMs perform position reasoning with segmentation, enabling an interactive framework that responds to contextual questions and generates both position reasoning outputs and corresponding segmentation masks.

\noindent\textbf{Multimodal-LLM in Medical Imaging.} Multimodal large language models (MLLMs) have recently shown promising results in medical image reasoning tasks. Prominent methods such as Med-Flamingo~\cite{medflamingo}, Med-MoE~\cite{medmoe}, GSCo~\cite{GSCo}, HuatuoGPT~\cite {chen2024huatuogpt}, and MedVLM-R1~\cite{medvlm-r1} are built upon vision and Language models like LLaVA~\cite{llava}, Qwen2-VL~\cite{qwen2vl}, and Multimodal Llama~\cite{llama} through the training technique via visual instruction tuning or reinforcement learning methods. Despite their promising results, these models struggle with medical image segmentation, which is a critical task for accurate disease diagnosis. Moreover, they also inherit spatial reasoning limitations from their Multimodal-LLMs, which are observed in SpatialVLM~\cite{spatialvlm}, Loc-VLM~\cite{locvlm}, and Spatialrgpt~\cite{spatialrgpt}. To address the spatial reasoning challenge, we propose PosMed, a position reasoning dataset to perform the localization understanding of the MLLMs in the medical domain. This represents an initial step toward spatial reasoning, and it can be extended for further grounding and spatial perception of the MLLMs in the medical domain.

\noindent\textbf{Reasoning Image Segmentation.}Recent advancements in reasoning segmentation have begun to integrate high-level reasoning, particularly through the use of the Multimodal-LLMs. Notable works in the domain include LISA~\cite{lisa}, LLM-Seg~\cite{wang2024llm}, and SegLLM~\cite{wang2024segllm}, which include a special [\texttt{SEG}] token used to compact segmentation-specific embeddings from the model. Although the initial success of previous works, they are still limited to the ability to perform single-object identification, and cannot perform long-form reasoning for segmentation. Moreover, recent MLLMs in the medical domain restrict the ability to diagnose, which is difficult to deal with, with more advanced reasoning abilities such as spatial reasoning and grounding ability.  To address this limitation, the PosMed dataset is designed for dealing with the limitations in the position reasoning segmentation. In addition, PRS-Med is a unified model with the reasoning and segmentation ability for position-aware reasoning, along with the segmentation visualization. Our contributions aim to enhance the ability of the MLLMs in spatial reasoning, along with the segmentation visualization for the interpretation, which is important in the clinician application.
\section{PosMed dataset}
\label{method:dataset}

\begin{figure*}[ht]
    \centering
    \begin{minipage}{0.48\linewidth}
        \centering
        \includegraphics[width=\linewidth]{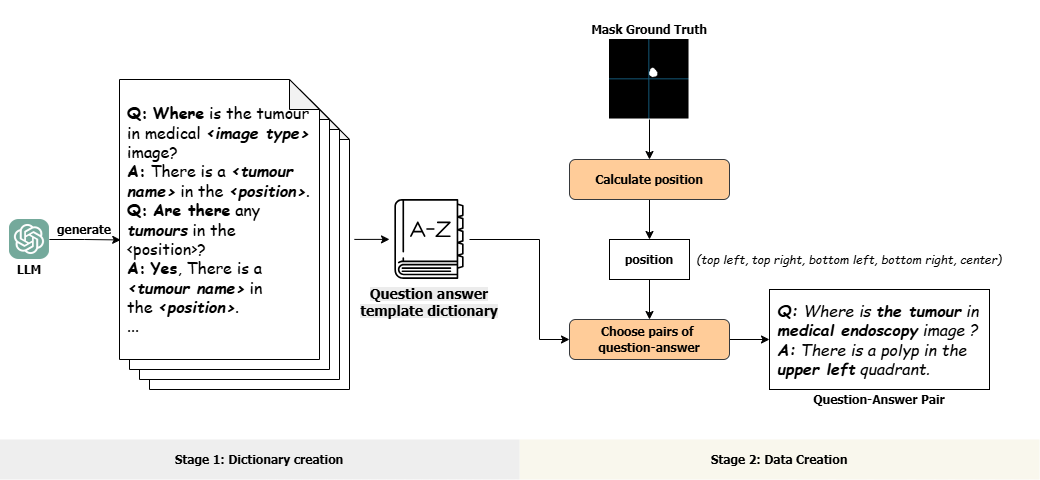}
        \caption{Visualization of two stages of the PosMed dataset pipeline.}
        \label{fig:dataset_creation_pipeline}
    \end{minipage}
    \hfill
    \begin{minipage}{0.48\linewidth}
        \centering
        \includegraphics[width=\linewidth]{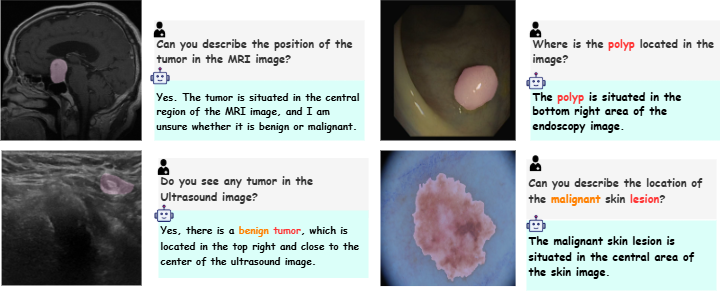}
        \caption{Visualization of the segmentation masks and question-answer pairs from the PosMed dataset.}
        \label{fig:dataset_vis}
    \end{minipage}
\end{figure*}

To address the challenge of the segmentation along with the position reasoning, we propose the PosMed dataset pipeline (presented in Figure~\ref{fig:dataset_creation_pipeline}), a two-stage dataset pipeline for position reasoning segmentation, which leverages the segmentation mask annotations to create the position annotation for the position reasoning, and by leveraging LLM collaborates with the doctor, we can semi-automate the dataset creation process. As a result, we release the PosMed dataset, a large-scale position reasoning segmentation.

\noindent\textbf{Data source.} We create the PosMed dataset by curating and extracting the spatial information from several datasources, including BUSI~\cite{busi} for breast; Brain MRI~\cite{braintumour1,braintumour2} for brain tumors in the MRI; LungCT dataset~\cite{lungct} for the lesions in the CT modality, and Lung Xray~\cite{lungxray1,lungxray2} for the lesion in the Xray modality; Kvasir-SEG~\cite{kvasir-seg}, ClinicDB~\cite{clinicdb}, CVC300~\cite{cvc300}, ColonDB~\cite{colondb}, and ETIS~\cite{etis} for the polyp in the endoscopy images; and ISIC~\cite{isic1} for the skin lesion in the RGB image domain. In overall, there are total of 38731 images in our dataset. These data sources ensure the diversity and the variety in the medical image modalities, anatomy types, and tumor types in the medical domain.

\noindent \textbf{Dataset Construction Pipeline.} To begin, we leverage the GPT-4 model to generate 55 different question-and-answer templates (pairs) based on the mentioned question-answer pair. All of the question and answer pairs always include these two pieces of information \emph{$<$tumor/anatomy name$>$}, and \emph{$<$position$>$}. These templates are then validated by three doctors to ensure the correctness in the medical context, and to ensure that when combined with the tumor name and positional information, the resulting sentences are coherent and contextually appropriate to provide the necessary information to the doctor. 

Then, to extract positional information from the segmentation mask follow the Algorithm~\ref{alg:extract_position}. Given a binary mask $X_{\text{mask}}$, we first derive the bounding box $\{x, y, w, h\}$ for the location of the tumor/anatomy. From this, we calculate the center point of the tumor as $x_{\text{center}} = \{x + \frac{w}{2}, y + \frac{h}{2}\}$. Next, we divide the image into anatomical action zone (AAZ) top left, top right, bottom left, and bottom right—as illustrated in Figure~\ref{fig:dataset_creation_pipeline}. The choice of a region extraction is a deliberate design decision to minimize spatial hallucination and maximize diagnostic reliability, which allows for the creation of massive, expert-validated datasets (116,000 pairs) without the noise inherent in manual free-form labeling.  Based on the location of $x_{\text{center}}$, we determine which AAZ region the tumor lies in and assign it a corresponding label. In addition to handling cases where tumors are located near the image center, we also compute the distance between $x_{\text{center}}$ and the geometric center of the image. If this distance falls below a predefined threshold $\tau$ pixel value, we label the tumor as being near the center.  

After getting the position information, we integrate the extracted positional information along with the tumor/anatomy type from the dataset with the question-and-answer templates to generate the final dataset of spatially grounded tumor descriptions. Afterward, we employ the GPT model to polish the question and answer pairs to mitigate the coarse problems in linguistics. Finally, we filter out the repeated question and answer pairs in each image to get about 310,086 pairs. The example of question and answer results is illustrated in Figure~\ref{fig:dataset_vis}. 

\begin{algorithm}[!ht]
\caption{Extract tumor/anatomy AAZ regions from a binary segmentation mask with $\{TL, TR, BL, BR\}$ representing Top Left, Top Right, Bottom Left, Bottom Right.}
\label{alg:extract_position}
\begin{algorithmic}[1]
\Require Image size $(H,W)$; binary mask $X_{\text{mask}} \in \{0,1\}^{H\times W}$; center threshold $\tau>0$
\Ensure $L \in \mathcal{L}$ (or \texttt{INVALID}), where $\mathcal{L}=\{\texttt{TL},\texttt{TR},\texttt{BL},\texttt{BR},\texttt{CENTER}\}$

\State $\Omega \gets \{(u,v)\mid X_{\text{mask}}(u,v)=1\}$  \Comment{$u$: row (y), $v$: column (x)}
\If{$\Omega = \emptyset$}
  \State \Return \texttt{INVALID}
\EndIf

\State $x_{\min}\gets \min_{(u,v)\in\Omega} v,\;\; x_{\max}\gets \max_{(u,v)\in\Omega} v$
\State $y_{\min}\gets \min_{(u,v)\in\Omega} u,\;\; y_{\max}\gets \max_{(u,v)\in\Omega} u$
\State $w \gets x_{\max}-x_{\min},\;\; h \gets y_{\max}-y_{\min}$

\State $x_c \gets x_{\min} + \frac{w}{2},\;\; y_c \gets y_{\min} + \frac{h}{2}$ \Comment{tumor center from bbox}
\State $x_I \gets \frac{W}{2},\;\; y_I \gets \frac{H}{2}$ \Comment{image center}

\State $d \gets \sqrt{(x_c-x_I)^2 + (y_c-y_I)^2}$
\If{$d \le \tau$}
  \State \Return \texttt{CENTER}
\EndIf

\If{$x_c < x_I \land y_c < y_I$}
  \State $L \gets \texttt{TL}$
\ElsIf{$x_c \ge x_I \land y_c < y_I$}
  \State $L \gets \texttt{TR}$
\ElsIf{$x_c < x_I \land y_c \ge y_I$}
  \State $L \gets \texttt{BL}$
\Else
  \State $L \gets \texttt{BR}$
\EndIf

\State \Return $L$

\end{algorithmic}
\end{algorithm}


\noindent\textbf{Expert-in-the-loop Validation.} 
To ensure the \textbf{(1)} \emph{medical accuracy} and \textbf{(2)} \emph{clinical demand} of the PosMed dataset, we adopt a validation process supported by three board-certified radiologists; each sample is identified as 0 for unqualified and 1 for qualified. Each radiologist reviews the generated question–answer pairs independently, focusing on the \emph{medical content} (correctness of anatomical or tumor descriptions), \emph{positional reasoning} (relevant to the anatomy or tumor annotations in the image), and \emph{clinical plausibility} (the question and answers are useful for radiologists). The final decision for each sample is obtained through majority voting, where only samples approved by the majority of experts are retained in the dataset. In addition, qualitative feedback from radiologists is used to refine the question-and-answer templates, thereby correcting ambiguous expressions, enhancing clinical clarity, and reducing potential biases or inconsistencies. This expert-driven validation framework ensures that PosMed maintains high fidelity in both medical semantics and spatial reasoning. Overall, after the validation stage, we obtain 116000 filtered question and answer pairs from 38731 images. Our dataset pipeline allows for multi-modal scaling (6 modalities) that would be impossible with manual spatial annotations.

\begin{figure}[!ht]
    \centering
    \begin{minipage}[t]{0.52\linewidth}
        \centering
        \includegraphics[width=\linewidth]{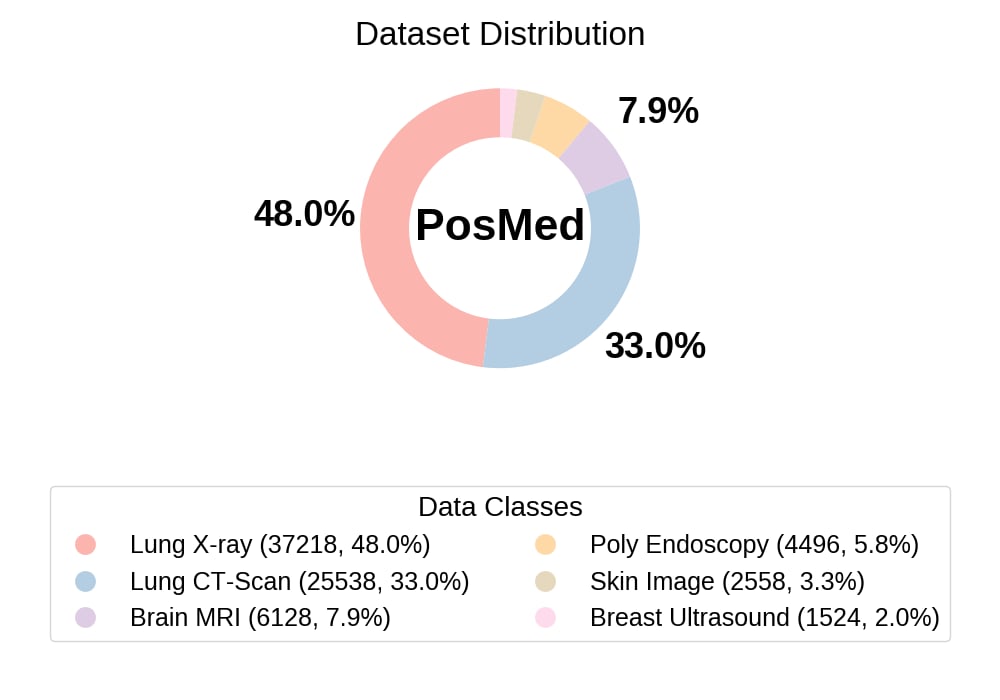}
        \label{fig:posmed_distribution}
    \end{minipage}\hfill
    \begin{minipage}[t]{0.48\linewidth}
        \centering
        \includegraphics[width=\linewidth]{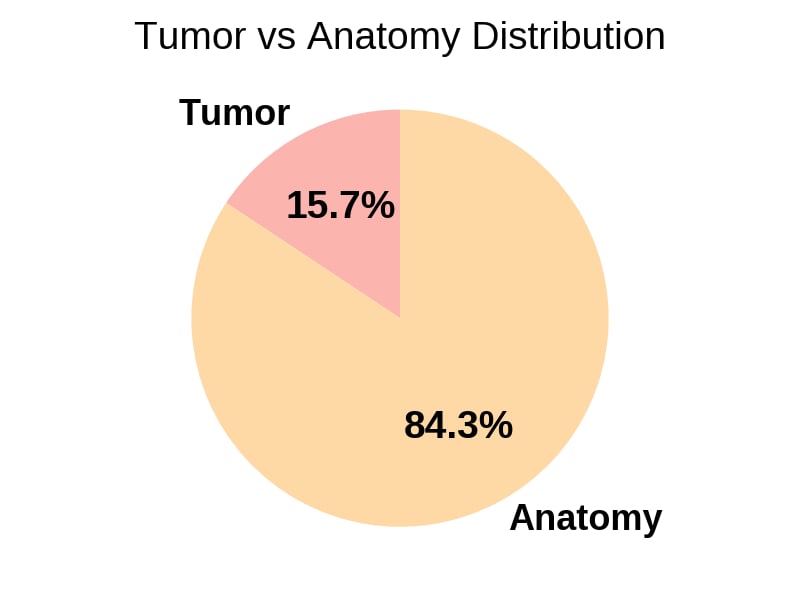}
        \label{fig:posmed_tumor_ana}
    \end{minipage}
    \caption{Statistical analysis of the PosMed. \textbf{Left:} Data-source distribution of six classes. \textbf{Right:}Tumor vs. anatomy composition.}
    \label{fig:statistic}
\end{figure}

\noindent\textbf{Dataset Statistic.} Figure~\ref{fig:statistic} summarizes the PosMed dataset distribution. Lung X-ray constitutes the largest portion (48.0\%), followed by lung CT (33.0\%) and Brain MRI (7.9\%). The remaining samples come from polyp endoscopy, skin images, and breast ultrasound, which together form a smaller share but increase the dataset diversity across multiple anatomies and tumor types. In terms of sample composition, PosMed is dominated by anatomy-centric descriptions (84.3\%), while tumor-centric samples account for 15.7\%.

\section{PRS-Med}
\label{method:PRS}
\begin{figure*}[ht]
    \centering
    \includegraphics[width=0.75\linewidth]{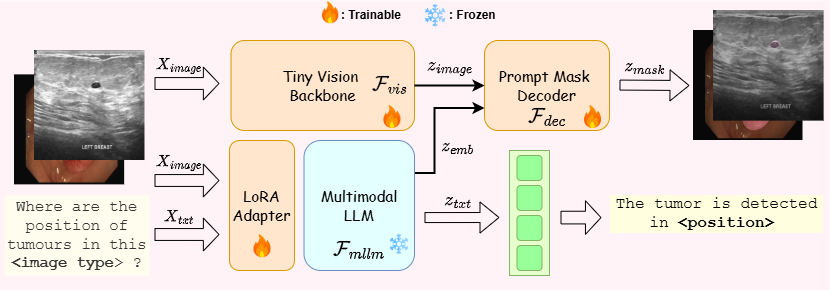}
    \caption{The architecture of PRS-Med comprises three primary components: (1) the Tiny Vision Backbone, (2) the Prompt Mask Decoder, and (3) the Multimodal-LLM. The framework accepts two input modalities: an image and a text-based prompt (e.g., a question). The image is processed through a vision encoder, while the prompt is embedded via a LoRA-adapted Multimodal-LLM. The fused representations are used to produce two outputs: a segmentation mask for the tumor regions, and a textual description specifying the tumor's location.}
    \label{fig:overall_arch}
\end{figure*}

\textbf{Overall Architecture.} The primary goal of PRS-Med is to perform position reasoning segmentation, enabling the model to explain the location of tumors or anatomies in an image along with relevant medical information. Additionally, the segmentation head allows the model to perform tumor segmentation within the image using a single prompt. The overall architecture is illustrated in Figure~\ref{fig:overall_arch}. 

    %
\noindent This framework consists of three main modules. The first is the Vision-Language Model, we employ LLaVA-Med \cite{llavamed}, as it is a well-trained Multimodal-LLM for the medical dataset. The second module is the Tiny SAM image encoder, employed from TinySAM \cite{tinysam}, which is used to encode the input image. The third module is our proposed Prompt Mask Decoder, which includes our proposed fusion component that combines image features from the image encoder with the vision-language embeddings from the Medical Vision-Language Model to generate the final segmentation mask. In addition, we include a Language Model Head to perform the reasoning task.

During training, due to the challenges of fine-tuning the full LLaVA-Med model, we apply Low-Rank Adaptation (LoRA) \cite{hu2022lora} to enable the model to effectively learn position reasoning information from our prepared dataset.

\noindent\textbf{Vision Backbone.} The primary objective of the vision backbone is to extract pixel-level features from medical images to support conditional segmentation. For this purpose, we adopt the image encoder from TinySAM \cite{tinysam}, which is based on the lightweight TinyViT architecture \cite{tinyvit}. This design enables efficient image encoding while reducing computational resource requirements, and adapt to the medical domain without initializing weights from scratch.

Given a batch of $b$ input images $X_{image} \in \mathbb{R}^{b \times 3 \times W \times H}$, the images are processed through a tiny vision transformer model $\mathcal{F}{vis}$, consisting of approximately four transformer layers, to produce an image representation embedding $z_{image} \in \mathbb{R}^{b \times 256 \times \frac{W}{16} \times \frac{H}{16}}$. The ablation for the choice of this TinyViT-based vision backbone is detailed in Section~\ref{sec:experiments}.



\noindent\textbf{Multimodal-LLM.} Most current Multimodal-LLM backbones applied to the medical domain—such as Flamingo \cite{flamingo}, LLaVA \cite{llava}, Qwen-VL \cite{qwen2vl}, and InternVL \cite{internvl}—demonstrate strong reasoning capabilities. However, they cannot generally generate masks for visual recognition tasks and struggle to comprehend positional information, such as the position of objects within medical images. In this work, we leverage the LLaVA-Med~\cite{llavamed}, which is trained on the extensive medical image dataset as the Language Model backbone. To allow the LLaVA-Med segmentation ability along with the reasoning, we leverage the LLaVA-Med's last hidden state embedding $z_{emb} \in \mathbb{R}^{b \times l \times 4096}$ (where $l$ is the token length) as the conditioning input for the segmentation head. In detail, we generate the $z_{emb}$, and the reasoning output $z_{txt} \in \mathbb{R}^{b \times l}$ from the input image $X_{image} \in \mathbb{R}^{b \times 3 \times w \times h}$ and input text $X_{txt} \in \mathbb{R}^{b \times l \times d}$ (where $d$ is the vocabulary size), we define $F_{mllm}$ as a parametric function. The autoregressive process of the model is described in Equation~\ref{equa:1} and Equation~\ref{equa:2}.



\begin{equation}
z_{emb} = \mathcal{F}_{mllm}(X_{image}, X_{txt}), 
\label{equa:1}
\end{equation}

\begin{equation}
\begin{aligned}
    z_{txt} = \prod_{i=1}^{l}p_{\theta}(z_{txt}^{i}|X_{image}, X_{txt}^{i-1})
    \label{equa:2}
\end{aligned}
\end{equation}

where $\theta$ is the trainable parameter. In our case, $\theta$ is from the parameter of the parametric function $F_{mllm}$.

During training, due to the high computational cost associated with fully supervised fine-tuning of the LLaVA-Med model, we employ the LoRA method \cite{hu2022lora} as an adapter. This approach enables the model to learn reasoning from our position-aware medical reasoning dataset while adapting to produce informative embeddings for the mask decoder, thereby mitigating catastrophic forgetting during multi-task MLLM training. The reason and LoRA hyperparameter choices are ablated in Section~\ref{sec:ablation}. By training the MLLM on clear spatial zones from PosMed, the model learns a strong spatial prior, which helps the Prompt Mask Decoder focus its attention, allowing it to capture small lesions that most grounded models miss.

\noindent\textbf{Prompt Mask Decoder.} The goal of this module is to predict the segmented mask from two inputs, including medical images representation feature $z_{image}$ and the embedded image-text prompt $z_{emb}$ from the Multimodal-LLM. This decoder module includes two parts: the fusion module and the mask prediction module. This design allows dynamic alignment between image regions and positional phrases, making better alignment between spatial features and medical vocabulary.


\noindent\textit{Fusion Module:} Given the image representation from the vision encoder, denoted as $z_{\text{image}} \in \mathbb{R}^{b \times 256 \times 16 \times 16}$, and the conditioning input from the Multimodal-LLM, denoted as $z_{\text{emb}} \in \mathbb{R}^{b \times l \times 4096}$, the overall fusion process is formalized in Equation~\ref{equa:cross-attn} and Equation~\ref{equa:skip_con}.

\begin{figure*}[t]
\centering
\begin{equation}
z_{fused} = MHA(\sigma(\frac{\mathcal{F}^{proj}_{\theta_{1}}(z_{image})\mathcal{F}^{proj}_{\theta_{2}}(z_{emb})^{T})}{\sqrt{d_{k}}})\mathcal{F}^{proj}_{\theta_{2}}(z_{emb})) ,
    \label{equa:cross-attn}
\end{equation}

\begin{equation}
    z_{fused} = z_{fused} + z_{image} .
    \label{equa:skip_con}
\end{equation}
\end{figure*}

\noindent where $d_{k}$ is the scaling value, $MHA(.)$ is the Multi-head Attention layers, and $\sigma(.)$ is the softmax function.

First, the image representation $z_{\text{image}}$ is reshaped to a new form $z_{\text{image}} \in \mathbb{R}^{b \times (16 \times 16) \times 256}$ to enable interaction with the embedding $z_{\text{emb}} \in \mathbb{R}^{b \times l \times 4096}$ from the Multimodal-LLM. As shown in Equation~\ref{equa:cross-attn}, two projection layers, $\mathcal{F}^{\text{proj}}_{\theta_1}$ and $\mathcal{F}^{\text{proj}}_{\theta_2}$, are applied to project both features into a shared latent space of dimension 256. This alignment allows effective fusion through a cross-attention mechanism, which integrates the image features with the Multimodal-LLM's embeddings. The choice of cross-attention is motivated by the dynamic length of the $z_{\text{emb}}$ sequences, making it a more flexible and suitable alternative to simple addition or concatenation. Following the fusion, a self-attention layer is employed to model the internal dependencies within the target sequence. The resulting fused representation, $z_{\text{fused}} \in \mathbb{R}^{b \times (16 \times 16) \times d}$, is then reshaped to $\mathbb{R}^{b \times 256 \times 16 \times 16}$. Finally, as described in Equation~\ref{equa:skip_con}, a skip connection is introduced to preserve gradient flow and mitigate the vanishing gradient problem during training. 

\noindent\textit{Mask Prediction Module:} The input $z_{\text{fused}}$ is passed through a stack of transposed 2D convolutional layers, each followed by Batch Normalization and ReLU activation. This series of operations progressively upsamples $z_{\text{fused}}$ to produce the final segmentation output $z_{\text{mask}} \in \mathbb{R}^{b \times 1 \times 1024 \times 1024}$.

\noindent\textbf{Objective Function.} This model is post-trained by using the segmentation loss ($\mathcal{L}_{seg}$) and text generation loss $\mathcal{L}_{text}$. The overall objective function is depicted in Equation~\ref{equa:overall_loss}.
\vspace{-1mm}
\begin{equation}
    \mathcal{L} = \lambda_{seg}\mathcal{L}_{seg} + \lambda_{txt}\mathcal{L}_{text}.
    \label{equa:overall_loss}
\end{equation}
where $\lambda_{seg}$ and $\lambda_{txt}$ show the importance of each loss in the overall framework. In our training setup, we set these values to 1 and 0.5, respectively.

Regarding $\mathcal{L}_{seg}$, we employ a combination of Binary Cross-Entropy and Dice loss \cite{diceloss}, which is a common choice in image segmentation tasks. For $\mathcal{L}_{txt}$, we use the Categorical Cross-Entropy (CE) loss applied on the logit vectors of the tokens output. Let $\hat{y}_{mask}$ denote the ground truth mask and $z_{mask}$ the predicted mask; similarly, let $\hat{y}_{txt}$ be the ground truth token index sequence and $z_{txt}$ the predicted text logits. Equations~\ref{equa:seg} and~\ref{equa:text} illustrate the formulations of the aforementioned loss functions $\mathcal{L}_{seg}$, and $\mathcal{L}_{txt}$.

 \begin{gather}
     \mathcal{L}_{seg} = \mathcal{L}_{BCE}(\hat{y}_{mask}, z_{mask})  + \mathcal{L}_{dice} (\hat{y}_{mask}, z_{mask})
     \label{equa:seg}
     \\
     \mathcal{L}_{text} = \mathcal{L}_{CE}(\hat{y}_{txt}, z_{txt}).
     \label{equa:text}
 \end{gather}

By employing this objective function, PRS-Med can simultaneously learn position reasoning while also learning to perform segmentation. Notably, during training, the decoder receives gradients not only from segmentation losses but also from textual reasoning losses, creating a feedback loop where segmentation informs reasoning and vice versa.

\section{Experimental Setup}
\label{sec:experiments}

\paragraph{Dataset.}
We train on the \textit{PosMed} dataset (consisting of six types of images: ultrasound, MRI, RGB image, CT Image, X-ray, and endoscopy), combining images from BUSI~\cite{busi}, BrainMRI~\cite{braintumour1, braintumour2}, ISIC~\cite{isic1}, LungCT~\cite{lungct}, LungXray~\cite{lungxray1, lungct}, Kvasir-SEG, and ClinicDB~\cite{clinicdb}. We follow the original train/test splits from each source dataset. To evaluate generalization on unseen distributions, we augment the endoscopy test set with CVC-300~\cite{cvc300}, ETIS~\cite{etis}, and ColonDB~\cite{colondb}, which are excluded from training.


\noindent\textbf{Implementation Details.}
All experiments are conducted on two NVIDIA H100 80GB GPUs. We train for 20 epochs with a batch size of 8 per GPU using AdamW~\cite{adamw} with a learning rate of $1 \times 10^{-4}$. For LoRA, we set rank $r = 16$, $\alpha = 16$, and dropout $= 0.05$, with weights initialized from a uniform distribution. LoRA is applied to the query, key, value, and output projection matrices of all attention layers in LLaVA-Med. The TinySAM vision encoder is fully unfrozen during training.


\noindent\textbf{Baselines.}
To compare our work with SOTA methods, following the reasoning segmentation task mentioned by LISA~\cite{lisa} and the diagnostic reasoning of medical MLLM~\cite{llavamed}, we conduct three benchmarks, including segmentation, position reasoning, and position understanding. For the Segmentation task, we compare our methods with the Foundation Segmentation model of medical imaging, such as SAM-Med 2D \cite{zhu2024medical} (2024), and Biomedparse \cite{biomedparse} (2024) (finetuned image encoder and decoder on our dataset), and the reasoning segmentation model, which is also finetuned on our dataset, is LISA \cite{lisa} with two versions are 7B and 13 B. Regarding the SAM model in medical imaging, there is a challenge that most medical segmentation model is based on the box prompt. For this reason, we leverage the Grounding Dino \cite{liu2024grounding} as the text understanding model to extract the boxes coordinates for the segmentation task. In the Position Reasoning benchmark, due to the lack of methods done reasoning segmentation, we reproduce the fine-tuning process on our dataset for the Multimodal-LLM for medical image, which includes LLaVA-Med \cite{llavamed} (2024), HuatuoGPT-Vision \cite{chen2024huatuogpt} (2024), Med-MoE \cite{medmoe}, and MedVLM-R1 \cite{medvlm-r1} (2025) to do the reasoning benchmark. In all of the comparisons, we do the fine-tuning of these methods on our PosMed dataset with the best-practice hyperparameter for each method for the fairest comparison.

\noindent\textbf{Evaluation Metric.} For evaluation, we use mDice and mIoU to benchmark segmentation performance, following standard practice in medical image segmentation. To assess the fluency and the accuracy of position reasoning, we report ROUGE score, and we use accuracy to measure the correctness of the model’s reasoning answers.
\section{Evaluation Results}
\label{sec:result}
\subsection{Quantitative Results}
\noindent\textbf{Segmentation Task Results.}
To evaluate the overall performance of PRS-Med in the segmentation task, we compare our method with prior works as aforementioned. Table~\ref{table:quality_full} presents results on radiology images of six different images and tissues, including Breast Ultrasound, Brain MRI, Lung CT-Scan, Lung X-ray, Polyp Endoscopy, and Skin Image.

\begin{table*}[t]
\centering
\setlength{\tabcolsep}{3pt}
\renewcommand{\arraystretch}{0.95}
\resizebox{\textwidth}{!}{%
\begin{tabular}{@{}lcccccccccccc@{}}
\toprule
\multirow{2}{*}{\textbf{Method}} 
& \multicolumn{2}{c}{\textbf{Breast Ultrasound}} 
& \multicolumn{2}{c}{\textbf{Brain MRI}} 
& \multicolumn{2}{c}{\textbf{Lung CT-Scan}} 
& \multicolumn{2}{c}{\textbf{Lung X-ray}} 
& \multicolumn{2}{c}{\textbf{Polyp Endoscopy}} 
& \multicolumn{2}{c}{\textbf{Skin Image}} \\
& \textbf{mDice} $\uparrow$ & \textbf{mIoU} $\uparrow$
& \textbf{mDice} $\uparrow$ & \textbf{mIoU} $\uparrow$
& \textbf{mDice} $\uparrow$ & \textbf{mIoU} $\uparrow$
& \textbf{mDice} $\uparrow$ & \textbf{mIoU} $\uparrow$
& \textbf{mDice} $\uparrow$ & \textbf{mIoU} $\uparrow$
& \textbf{mDice} $\uparrow$ & \textbf{mIoU} $\uparrow$ \\
\midrule
G\mbox{-}Dino + SAM\mbox{-}Med2D~\cite{ma2024segment} 
& 0.515 & 0.441 & \underline{0.667} & \underline{0.625} & 0.540 & 0.392 & 0.401 & 0.300 & 0.488 & 0.418 & 0.237 & 0.171 \\
Biomedparse~\cite{biomedparse} 
& \underline{0.783} & \underline{0.698} & 0.294 & 0.245 & 0.516 & 0.399 & \underline{0.972} & \underline{0.949} & \underline{0.824} & \underline{0.774} & \underline{0.893} & \underline{0.822} \\
LISA\mbox{-}7B~\cite{lisa} 
& 0.299 & 0.246 & 0.478 & 0.402 & 0.478 & 0.402 & 0.397 & 0.263 & 0.241 & 0.202 & 0.464 & 0.368 \\
LISA\mbox{-}13B~\cite{lisa} 
& 0.705 & 0.680 & 0.439 & 0.357 & \underline{0.656} & \underline{0.528} & 0.664 & 0.535 & 0.312 & 0.247 & 0.643 & 0.536 \\
\midrule
\textbf{PRS\mbox{-}Med} 
& \textbf{0.817} & \textbf{0.729} & \textbf{0.803} & \textbf{0.757} & \textbf{0.968} & \textbf{0.943} & \textbf{0.973} & \textbf{0.952} & \textbf{0.843} & \textbf{0.791} & \textbf{0.901} & \textbf{0.833} \\
\textit{vs previous works} 
& \textbf{\textcolor{teal}{+0.034}} & \textbf{\textcolor{teal}{+0.031}} 
& \textbf{\textcolor{teal}{+0.136}} & \textbf{\textcolor{teal}{+0.132}} 
& \textbf{\textcolor{teal}{+0.312}} & \textbf{\textcolor{teal}{+0.415}} 
& \textbf{\textcolor{teal}{+0.001}} & \textbf{\textcolor{teal}{+0.002}}
& \textbf{\textcolor{teal}{+0.019}} & \textbf{\textcolor{teal}{+0.017}}
& \textbf{\textcolor{teal}{+0.008}} & \textbf{\textcolor{teal}{+0.011}} \\
\bottomrule
\end{tabular}%
}
\caption{Quantitative results of PRS-Med across six medical image types. The highest score in each column is in \textbf{bold}; the second highest is \underline{underlined}. PRS-Med shows its competitive results when it surpasses all of the previous works across six datasets.}
\label{table:quality_full}
\end{table*}

\noindent As shown in Table~\ref{table:quality_full}, PRS-Med achieves competitive results with state-of-the-art. To the second-best method, the improvements (mDice, mIoU) are $(+3.4\%, +3.1\%)$ on Breast Ultrasound, $(+13.6\%, +13.2\%)$ on Brain MRI, $(+31.2\%, +41.5\%)$ on Lung CT-Scan, $(+0.1\%, +0.2\%)$ on Lung X-ray, $(+1.9\%, +1.7\%)$ on Polyp Endoscopy, and $(+0.8\%, +1.1\%)$ on Skin Images. These results highlight the generalization and robustness of PRS-Med across imaging modalities, anatomical structures, and tumors.

\noindent\textbf{Position Reasoning Results.}
To assess the performance of PRS-Med, we evaluate the output of linguistic content with coherence and the accuracy of position reasoning with SOTA methods in Multimodal-LLM for medical images, which is shown in Table~\ref{tab:prsmed_reasoning}. 

\begin{table*}[!ht]
\centering
\setlength{\tabcolsep}{3.0pt}
\renewcommand{\arraystretch}{0.95}
\resizebox{\textwidth}{!}{%
\begin{tabular}{@{}l*{12}{c}@{}}
\toprule
\multirow{2}{*}{\textbf{Method}} 
& \multicolumn{2}{c}{\textbf{Breast Ultrasound}}    
& \multicolumn{2}{c}{\textbf{Brain MRI}}     
& \multicolumn{2}{c}{\textbf{Lung CT-Scan}}      
& \multicolumn{2}{c}{\textbf{Lung X-ray}}     
& \multicolumn{2}{c}{\textbf{Polyp}}     
& \multicolumn{2}{c}{\textbf{Skin Image}} \\
& \textbf{ROUGE} $\uparrow$ & \textbf{ACC} $\uparrow$
& \textbf{ROUGE} $\uparrow$ & \textbf{ACC} $\uparrow$
& \textbf{ROUGE} $\uparrow$ & \textbf{ACC} $\uparrow$
& \textbf{ROUGE} $\uparrow$ & \textbf{ACC} $\uparrow$
& \textbf{ROUGE} $\uparrow$ & \textbf{ACC} $\uparrow$
& \textbf{ROUGE} $\uparrow$ & \textbf{ACC} $\uparrow$ \\
\midrule
LlaVA-Med \cite{llavamed}
& 0.33 & 75.22
& 0.33 & \underline{81.17}
& 0.32 & 74.17
& 0.33 & 81.09
& 0.30 & 53.76
& 0.29 & \underline{92.08} \\
HuatuoGPT \cite{chen2024huatuogpt}
& 0.36 & 79.65
& 0.36 & 25.50
& 0.35 & \textbf{88.17}
& 0.36 & 10.97
& 0.30 & \underline{55.64}
& 0.30 & 60.69 \\
Med-MoE \cite{medmoe}
& \underline{0.61} & \underline{81.63}
& \underline{0.66} & 67.00
& \underline{0.69} & 12.50
& \underline{0.61} & \underline{84.00}
& \underline{0.67} & 27.60
& \underline{0.68} & 81.58 \\
Med-VLMR1 \cite{medvlm-r1}
& 0.28 & 50.44
& 0.28 & 12.50
& 0.27 & 72.22
& 0.28 & 3.13
& 0.25 & 32.58
& 0.24 & 13.98 \\
\midrule
\textbf{PRS-Med}
& \textbf{0.64} & \textbf{92.92}
& \textbf{0.67} & \textbf{86.17}
& \textbf{0.71} & \underline{76.67}
& \textbf{0.64} & \textbf{94.36}
& \textbf{0.71} & \textbf{72.43}
& \textbf{0.76} & \textbf{96.31} \\
\textit{vs previous works}
& \textbf{\textcolor{teal}{+0.03}} & \textbf{\textcolor{teal}{+11.29}}
& \textbf{\textcolor{teal}{+0.01}} & \textbf{\textcolor{teal}{+5.00}}
& \textbf{\textcolor{teal}{+0.02}} & \textbf{\textcolor{orange}{-11.50}}
& \textbf{\textcolor{teal}{+0.03}} & \textbf{\textcolor{teal}{+10.36}}
& \textbf{\textcolor{teal}{+0.04}} & \textbf{\textcolor{teal}{+16.79}}
& \textbf{\textcolor{teal}{+0.08}} & \textbf{\textcolor{teal}{+4.23}} \\
\bottomrule
\end{tabular}%
}
\caption{Quantitative results of PRS-Med on the reasoning task across six medical image types. The highest score in each column is in \textbf{bold}, the second highest is \underline{underlined}.}
\label{tab:prsmed_reasoning}
\end{table*}

\noindent As shown in Table~\ref{tab:prsmed_reasoning}, PRS-Med achieves the best ROUGE across all six datasets and obtains the highest accuracy on five out of six datasets. Compared with the strongest prior baseline (Med\mbox{-}MoE), the absolute gains $(\text{ROUGE},\text{ACC})$ are: Breast Ultrasound $(+0.03,+11.29)$, Brain MRI $(+0.01,+5.00)$, Lung CT\mbox{-}Scan $(+0.02,-11.50)$, Lung X\mbox{-}ray $(+0.03,+10.36)$, Polyp $(+0.04,+16.79)$, and Skin Image $(+0.08,+4.23)$. Overall, PRS-Med consistently improves generation quality (ROUGE) while also strengthening task correctness (ACC), with the only exception on Lung CT\mbox{-}Scan where PRS-Med attains the best ROUGE but trails the accuracy of HuoGPT~\cite{chen2024huatuogpt}. Notably, although PRS-Med and LlaVA\mbox{-}Med~\cite{llavamed} share the same multimodal-LLM pretraining, PRS-Med yields substantially better results on position reasoning. We attribute this improvement to the jointly trained segmentation module, which injects explicit localization supervision during training and allows the LoRA adapters to better specialize for position-sensitive reasoning by updating their weights accordingly. These results also indicate that, during the learning of the segmentation, the MLLM does not forget its reasoning ability, which leads to the competitive performance with SOTA.

\subsection{Qualitative Analysis}
\begin{figure}[!ht]
    \centering
    \includegraphics[width=0.9\linewidth]{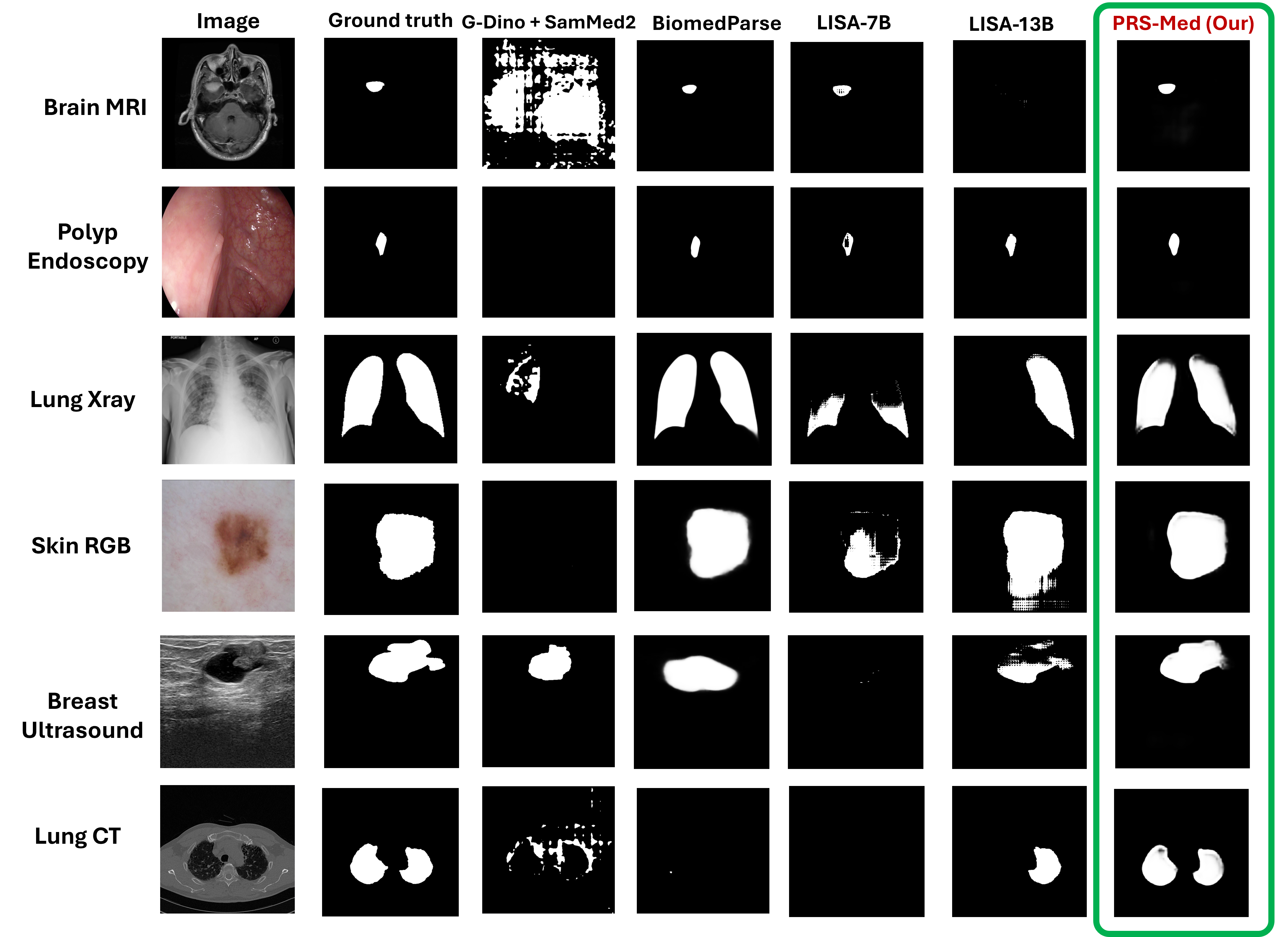}
    \caption{Comparison of PRS-Med with previous works. PRS-Med produces more accurate boundaries and captures small lesions missed by other methods.}
    \label{fig:qualitative}
\end{figure}

In Figure~\ref{fig:qualitative}, we present qualitative visualizations that highlight the improvements achieved by PRS-Med. The results clearly show that PRS-Med can capture small lesions and anatomies that previous baselines miss, consistently generating completed masks with the lowest loss. We attribute these improvements to the informative feature extraction of the lightweight vision encoder and the effectiveness of the fusion module. Overall, the results provide strong evidence for the promise of our approach. In addition, it shows that the discrete spatial grounding provided by PRS-Med acts as a 'visual prompt' that guides the decoder to identify small, high-stakes pathologies that generalist reasoning models overlook. However, through this visualization, we observe that the boundary problems are still the limitations of PRS-Med, and we are planning to improve in the future.

\vspace{-3mm}
\subsection{Ablation Study}
\label{sec:ablation}
To assess the choice and the effectiveness of the module in our framework, we conduct several experiments to assess the performance and the limitations of each module. The experiments are conducted in the same training and testing dataset with the benchmark. Regarding the metrics, we calculate the average mDice and average mIoU for the segmentation results, and ROUGE and METEOR for reasoning results (to assess the coherence in reasoning results) on different modalities in our test dataset to have the best assessment of the robustness of each choice. 

\begin{table}[!ht]
    \centering
    \begin{tabular}{cccc}
    \hline
        \textbf{Vision Backbone} & \textbf{Param} & \textbf{mDice} $\uparrow$ & \textbf{mIoU} $\uparrow$\\
        \hline
       SAM-Med (Frozen) & 21.52M & 0.798 & 0.719\\
       SAM-Med (Full) & 292.60M & 0.891 & 0.838 \\
       SAM-Med (LoRA) & 47.84M & 0.790 & 0.711 \\
       TinySAM (no pretrained) & 31.49M & 0.674  & 0.582 \\
       TinySAM (frozen) & 21.73M & 0.737 & 0.662 \\
       \textbf{TinySAM (Full)} & 31.49M & \textbf{0.884} & \textbf{0.834} \\
    \hline
    \end{tabular}
    \caption{Comparison results of different vision encoder backbones.}
    \label{tab:vision}
\end{table}

\begin{table}[!ht]
    \centering
    \begin{tabular}{ccccc}
        \hline
        \textbf{MLLM} & \textbf{mDice} $\uparrow$ & \textbf{mIoU} $\uparrow$ & \textbf{ROUGE} $\uparrow$ & \textbf{METEOR} $\uparrow$ \\
        \hline
       q,v & 0.573  & 0.483 & 0.478 & 0.436 \\
       q, k, v & 0.714 & 0.621 & 0.585 & 0.578 \\
       \textbf{q,k,v,o}  & \textbf{0.879} & \textbf{0.827} &  \textbf{0.654} & \textbf{0.599} \\
       \hline
    \end{tabular}
    \caption{Ablation study on LoRA target module (r$=$16) on Multimodal-LLM.}
    \label{tab:vision_lora_combined}
\end{table}

\noindent\textbf{Design Choice Of Vision Encoder Backbone.} As described in Section~\ref{method:PRS}, we assess the choice of vision encoder with the results presented in Table~\ref{tab:vision_lora_combined}. In our design, we consider two published pre-trained models, SAM-Med \cite{sammed2d} and TinySAM \cite{tinysam}, as our vision encoder. As demonstrated in Table~\ref{tab:vision_lora_combined}, TinySAM gets higher performance than SAM-Med (LoRA) with similar trainable parameters in the general framework and gets lower results with the full supervised training version of SAM-Med. Given the trade-off between efficiency and model accuracy, TinySAM is chosen as our vision encoder.

\noindent\textbf{Initialization Of TinySAM Image Encoder.} We observe that the initialization of TinySAM significantly affects the overall results. For this reason, in Table~\ref{tab:vision}, we assess the contribution of the TinySAM pretrained weight. Without the pretrained initialization, the overall results drop substantially, which shows the importance of the pretrained initialization to the overall framework.

\noindent\textbf{Design Choice Of MLLM Backbone.}
To evaluate the choice of MLLM backbone for PRS-Med, we conducted experiments comparing three models: LLaVA-1.5 \cite{liu2023improvedllava}, LLaVA-1.6 \cite{liu2024llavanext}, and LLaVA-Med~\cite{llavamed} using the same 7B backbone and fine-tuned via the LoRA approach. The comparisons are on two tasks: segmentation and position reasoning (shown in Table~\ref{tab:mllm_compare}). Results indicate that the performance of the LLaVA-Med baseline surpasses the remaining. This improvement can be attributed to LLaVA-Med’s enhanced adaptation to the medical domain, which enables it to better handle tasks involving medical data.


\begin{table}[!ht]
    \centering
    \small
    \setlength{\tabcolsep}{4pt}
    \renewcommand{\arraystretch}{1.15}
    \resizebox{\linewidth}{!}{%
    \begin{tabular}{lccc}
        \toprule
        \textbf{Metric} & \textbf{LLaVA-1.5 (LoRA)} & \textbf{LLaVA-1.6 (LoRA)} & \textbf{LLaVA-Med (LoRA)} \\
        \midrule
        \textbf{Param} & 34.63M & 31.49M & \textbf{31.49M} \\
        \textbf{Avg-mDice} $\uparrow$ & 0.709 & 0.744 & \textbf{0.879} \\
        \textbf{Avg-mIoU} $\uparrow$  & 0.642 & 0.671 & \textbf{0.827} \\
        \textbf{ROUGE} $\uparrow$     & 0.414 & 0.508 & \textbf{0.654} \\
        \textbf{METEOR} $\uparrow$    & 0.385 & 0.432 & \textbf{0.599} \\
        \bottomrule
    \end{tabular}%
    }
    \caption{Ablation study for design choice of the Multimodal-LLM.}
    \label{tab:mllm_compare}
\end{table}

\noindent\textbf{LoRA Target Modules Choice.} To assess the contribution of the target module from LoRA in the overall framework, we do the ablation to evaluate our choice of target module from LoRA, which is mentioned in Table~\ref{tab:vision_lora_combined}. Our choice for the target modules (q,k,v,o) makes the overall framework achieve significantly higher performances, which indicates that all of the projection weights allow LoRA to more effectively align cross-modal representations for both the reasoning and segmentation tasks.

\vspace{-2mm}
\section{Conclusion }
\label{sec:conclusion}
\vspace{-2mm}

In conclusion, we introduce \textbf{PosMed}, a large-scale dataset designed to benchmark and advance position-reasoning segmentation in medical imaging, together with \textbf{PRS-Med}, a strong baseline tailored to this challenge. Our experiments reveal two key gaps in current approaches: (i) text-referring segmentation methods remain brittle when the target is specified through spatial reasoning rather than direct textual cues, and (ii) existing medical MLLMs still struggle with position reasoning, a core capability for reliable perception in clinical settings. Importantly, results show that training and evaluating on PosMed, together with the PRS-Med framework, effectively addresses these limitations and yields competitive performance on position-reasoning segmentation. PRS-Med eschews the unnecessary complexity of pixel-coordinate regression in favor of a Categorical Spatial Grounding framework. This design choice ensures that the model’s reasoning is not only technically sound but clinically actionable (simple), providing a direct bridge between visual evidence and the structured language of medical diagnosis.  By releasing the PosMed data pipeline, model, and code, we hope to facilitate future work on spatially grounded multimodal reasoning and accelerate progress toward practical clinical applications.

\noindent\textbf{Future work.} We plan to extend PosMed and PRS-Med toward richer spatial reasoning beyond localization, enabling the model to identify relevant anatomy and tumors while also estimating clinically meaningful attributes such as lesion size and inter-structure distances. 

\noindent\textbf{Acknowledgement.} U. Bagci is supported partially by NIH R01 CA268808 and R01 HL171376. D. Jha is supported in part by the U.S. Department of Education (P116Z240151 to the University of South Dakota and the South Dakota School of Mines \& Technology). The views expressed are those of the author(s) and do not necessarily represent the official views of the U.S. Department of Education.  We would like to thank to Gorkem Durak, and Halil Ertugrul Aktas to support in reviewing the quality of the dataset.


{
    \small
    \bibliographystyle{ieeenat_fullname}
    \bibliography{main}

@String(ICLR = {Int. Conf. Learn. Represent.})

@String(AAAI = {AAAI})

@String(ICLR  = {ICLR})

@article{llavamed,
  title={Llava-med: Training a large language-and-vision assistant for biomedicine in one day},
  author={Li, Chunyuan and Wong, Cliff and Zhang, Sheng and Usuyama, Naoto and Liu, Haotian and Yang, Jianwei and Naumann, Tristan and Poon, Hoifung and Gao, Jianfeng},
  journal={Advances in Neural Information Processing Systems},
  volume={36},
  pages={28541--28564},
  year={2023}
}

@article{adamw,
  title={Decoupled weight decay regularization},
  author={Loshchilov, Ilya and Hutter, Frank},
  journal={arXiv preprint arXiv:1711.05101},
  year={2017}
}

@article{sammed2d,
  title={Sa-med2d-20m dataset: Segment anything in 2d medical imaging with 20 million masks},
  author={Ye, Jin and Cheng, Junlong and Chen, Jianpin and Deng, Zhongying and Li, Tianbin and Wang, Haoyu and Su, Yanzhou and Huang, Ziyan and Chen, Jilong and Jiang, Lei and others},
  journal={arXiv preprint arXiv:2311.11969},
  year={2023}
}

@article{medmoe,
  title={Med-moe: Mixture of domain-specific experts for lightweight medical vision-language models},
  author={Jiang, Songtao and Zheng, Tuo and Zhang, Yan and Jin, Yeying and Yuan, Li and Liu, Zuozhu},
  journal={arXiv preprint arXiv:2404.10237},
  year={2024}
}

@inproceedings{medflamingo,
  title={Med-flamingo: a multimodal medical few-shot learner},
  author={Moor, Michael and Huang, Qian and Wu, Shirley and Yasunaga, Michihiro and Dalmia, Yash and Leskovec, Jure and Zakka, Cyril and Reis, Eduardo Pontes and Rajpurkar, Pranav},
  booktitle={Machine Learning for Health (ML4H)},
  pages={353--367},
  year={2023}
}

@inproceedings{bui2024meganet,
  title={Meganet: Multi-scale edge-guided attention network for weak boundary polyp segmentation},
  author={Bui, Nhat-Tan and Hoang, Dinh-Hieu and Nguyen, Quang-Thuc and Tran, Minh-Triet and Le, Ngan},
  booktitle={Proceedings of the IEEE/CVF winter conference on applications of computer vision},
  pages={7985--7994},
  year={2024}
}

@inproceedings{locvlm,
  title={Learning to localize objects improves spatial reasoning in visual-llms},
  author={Ranasinghe, Kanchana and Shukla, Satya Narayan and Poursaeed, Omid and Ryoo, Michael S and Lin, Tsung-Yu},
  booktitle={Proceedings of the IEEE/CVF Conference on Computer Vision and Pattern Recognition},
  pages={12977--12987},
  year={2024}
}

@inproceedings{lisa,
  title={Lisa: Reasoning segmentation via large language model},
  author={Lai, Xin and Tian, Zhuotao and Chen, Yukang and Li, Yanwei and Yuan, Yuhui and Liu, Shu and Jia, Jiaya},
  booktitle={Proceedings of the IEEE/CVF Conference on Computer Vision and Pattern Recognition},
  pages={9579--9589},
  year={2024}
}

@inproceedings{spatialvlm,
  title={Spatialvlm: Endowing vision-language models with spatial reasoning capabilities},
  author={Chen, Boyuan and Xu, Zhuo and Kirmani, Sean and Ichter, Brain and Sadigh, Dorsa and Guibas, Leonidas and Xia, Fei},
  booktitle={Proceedings of the IEEE/CVF Conference on Computer Vision and Pattern Recognition},
  pages={14455--14465},
  year={2024}
}

@article{spatialrgpt,
  title={SpatialRGPT: Grounded Spatial Reasoning in Vision Language Models},
  author={Cheng, An-Chieh and Yin, Hongxu and Fu, Yang and Guo, Qiushan and Yang, Ruihan and Kautz, Jan and Wang, Xiaolong and Liu, Sifei},
  journal={arXiv preprint arXiv:2406.01584},
  year={2024}
}

@inproceedings{llm-seg,
  title={Llm-seg: Bridging image segmentation and large language model reasoning},
  author={Wang, Junchi and Ke, Lei},
  booktitle={Proceedings of the IEEE/CVF Conference on Computer Vision and Pattern Recognition},
  pages={1765--1774},
  year={2024}
}

@inproceedings{tinysam,
  title={Tinysam: Pushing the envelope for efficient segment anything model},
  author={Shu, Han and Li, Wenshuo and Tang, Yehui and Zhang, Yiman and Chen, Yihao and Li, Houqiang and Wang, Yunhe and Chen, Xinghao},
  booktitle={Proceedings of the AAAI Conference on Artificial Intelligence},
  volume={39},
  number={19},
  pages={20470--20478},
  year={2025}
}

@inproceedings{tinyvit,
  title={Tinyvit: Fast pretraining distillation for small vision transformers},
  author={Wu, Kan and Zhang, Jinnian and Peng, Houwen and Liu, Mengchen and Xiao, Bin and Fu, Jianlong and Yuan, Lu},
  booktitle={European conference on computer vision},
  pages={68--85},
  year={2022}
}

@inproceedings{diceloss,
  title={Generalised dice overlap as a deep learning loss function for highly unbalanced segmentations},
  author={Sudre, Carole H and Li, Wenqi and Vercauteren, Tom and Ourselin, Sebastien and Jorge Cardoso, M},
  booktitle={Deep Learning in Medical Image Analysis and Multimodal Learning for Clinical Decision Support: Third International Workshop, DLMIA 2017, and 7th International Workshop, ML-CDS 2017, Held in Conjunction with MICCAI 2017, Qu{\'e}bec City, QC, Canada, September 14, Proceedings 3},
  pages={240--248},
  year={2017}
}

@article{zhu2024medical,
  title={Medical sam 2: Segment medical images as video via segment anything model 2},
  author={Zhu, Jiayuan and Hamdi, Abdullah and Qi, Yunli and Jin, Yueming and Wu, Junde},
  journal={arXiv preprint arXiv:2408.00874},
  year={2024}
}

@article{llava,
  title={Visual instruction tuning},
  author={Liu, Haotian and Li, Chunyuan and Wu, Qingyang and Lee, Yong Jae},
  journal={Advances in neural information processing systems},
  volume={36},
  pages={34892--34916},
  year={2023}
}

@inproceedings{internvl,
  title={Internvl: Scaling up vision foundation models and aligning for generic visual-linguistic tasks},
  author={Chen, Zhe and Wu, Jiannan and Wang, Wenhai and Su, Weijie and Chen, Guo and Xing, Sen and Zhong, Muyan and Zhang, Qinglong and Zhu, Xizhou and Lu, Lewei and others},
  booktitle={Proceedings of the IEEE/CVF conference on computer vision and pattern recognition},
  pages={24185--24198},
  year={2024}
}

@article{flamingo,
  title={Flamingo: a visual language model for few-shot learning},
  author={Alayrac, Jean-Baptiste and Donahue, Jeff and Luc, Pauline and Miech, Antoine and Barr, Iain and Hasson, Yana and Lenc, Karel and Mensch, Arthur and Millican, Katherine and Reynolds, Malcolm and others},
  journal={Advances in neural information processing systems},
  volume={35},
  pages={23716--23736},
  year={2022}
}

@article{hu2022lora,
  title={Lora: Low-rank adaptation of large language models.},
  author={Hu, Edward J and Shen, Yelong and Wallis, Phillip and Allen-Zhu, Zeyuan and Li, Yuanzhi and Wang, Shean and Wang, Lu and Chen, Weizhu and others},
  journal={ICLR},
  volume={1},
  number={2},
  pages={3},
  year={2022}
}

@article{busi,
  title={Dataset of breast ultrasound images},
  author={Al-Dhabyani, Walid and Gomaa, Mohammed and Khaled, Hussien and Fahmy, Aly},
  journal={Data in brief},
  volume={28},
  pages={104863},
  year={2020}
}

@article{braintumour1,
  title={Enhanced performance of brain tumor classification via tumor region augmentation and partition},
  author={Cheng, Jun and Huang, Wei and Cao, Shuangliang and Yang, Ru and Yang, Wei and Yun, Zhaoqiang and Wang, Zhijian and Feng, Qianjin},
  journal={PloS one},
  volume={10},
  number={10},
  pages={e0140381},
  year={2015}
}

@article{braintumour2,
  title={Retrieval of brain tumors by adaptive spatial pooling and fisher vector representation},
  author={Cheng, Jun and Yang, Wei and Huang, Meiyan and Huang, Wei and Jiang, Jun and Zhou, Yujia and Yang, Ru and Zhao, Jie and Feng, Yanqiu and Feng, Qianjin and others},
  journal={PloS one},
  volume={11},
  number={6},
  pages={e0157112},
  year={2016}
}

@inproceedings{kvasir-seg,
  title={{Kvasir-SEG: A Segmented Polyp Dataset}},
  author={Jha, Debesh and Smedsrud, Pia H and Riegler, Michael A and Halvorsen, P{\aa}l and Lange, Thomas de and Johansen, Dag and Johansen, H{\aa}vard D},
  booktitle={Multimedia Modeling},
  year={2020}}

@article{clinicdb,
author = {Jorge Bernal and F. Javier Sánchez and Gloria Fernández-Esparrach and Debora Gil and Cristina Rodríguez and Fernando Vilariño},
title = {{WM-DOVA maps for accurate polyp highlighting in colonoscopy: Validation vs. saliency maps from physicians}},
journal = {CMIG},
pages = {99-111},
year = {2015}}

@ARTICLE{colondb,
  author={Tajbakhsh, Nima and Gurudu, Suryakanth R. and Liang, Jianming},
  journal={TMI}, 
  title={{Automated Polyp Detection in Colonoscopy Videos Using Shape and Context Information}}, 
  year={2016},
   pages={630-644}}

@article{etis,
  title = {{Towards embedded detection of polyps in WCE images for early diagnosis of colorectal cancer}},
  author = {Silva, Juan S. and Histace, Aymeric and Romain, Olivier and Dray, Xavier and Granado, Bertrand},
  journal = {IJCARS},
  PAGES = {283-293},
  YEAR = {2014}}

@article{cvc300,
  title={{A Benchmark for Endoluminal Scene Segmentation of Colonoscopy Images}},
  author={David V{\'a}zquez and Jorge Bernal and Francisco Javier S{\'a}nchez and Gl{\`o}ria Fern{\'a}ndez-Esparrach and Antonio M. L{\'o}pez and Adriana Romero and Michal Drozdzal and Aaron C. Courville},
  journal={Journal of Healthcare Engineering},
  year={2017}}

@article{lungct,
  title={Data from lung CT segmentation challenge (LCTSC)(version 3)[data set]},
  author={Yang, J and Sharp, G and Veeraraghavan, H and Van Elmpt, W and Dekker, A and Lustberg, T and Gooding, M},
  journal={The Cancer Imaging Archive},
  year={2017}
}

@article{lungxray1,
  title={Can AI help in screening viral and COVID-19 pneumonia?},
  author={Chowdhury, Muhammad EH and Rahman, Tawsifur and Khandakar, Amith and Mazhar, Rashid and Kadir, Muhammad Abdul and Mahbub, Zaid Bin and Islam, Khandakar Reajul and Khan, Muhammad Salman and Iqbal, Atif and Al Emadi, Nasser and others},
  journal={Ieee Access},
  volume={8},
  pages={132665--132676},
  year={2020}
}

@article{lungxray2,
  title={Exploring the effect of image enhancement techniques on COVID-19 detection using chest X-ray images},
  author={Rahman, Tawsifur and Khandakar, Amith and Qiblawey, Yazan and Tahir, Anas and Kiranyaz, Serkan and Kashem, Saad Bin Abul and Islam, Mohammad Tariqul and Al Maadeed, Somaya and Zughaier, Susu M and Khan, Muhammad Salman and others},
  journal={Computers in biology and medicine},
  volume={132},
  pages={104319},
  year={2021}
}

@inproceedings{isic1,
  title={Skin lesion analysis toward melanoma detection: A challenge at the 2017 international symposium on biomedical imaging (isbi), hosted by the international skin imaging collaboration (isic)},
  author={Codella, Noel CF and Gutman, David and Celebi, M Emre and Helba, Brian and Marchetti, Michael A and Dusza, Stephen W and Kalloo, Aadi and Liopyris, Konstantinos and Mishra, Nabin and Kittler, Harald and others},
  booktitle={2018 IEEE 15th international symposium on biomedical imaging (ISBI 2018)},
  pages={168--172},
  year={2018}
}

@article{medvlm-r1,
  title={Medvlm-r1: Incentivizing medical reasoning capability of vision-language models (vlms) via reinforcement learning},
  author={Pan, Jiazhen and Liu, Che and Wu, Junde and Liu, Fenglin and Zhu, Jiayuan and Li, Hongwei Bran and Chen, Chen and Ouyang, Cheng and Rueckert, Daniel},
  journal={arXiv preprint arXiv:2502.19634},
  year={2025}
}

@article{biomedparse,
  title={BiomedParse: a biomedical foundation model for image parsing of everything everywhere all at once},
  author={Zhao, Theodore and Gu, Yu and Yang, Jianwei and Usuyama, Naoto and Lee, Ho Hin and Naumann, Tristan and Gao, Jianfeng and Crabtree, Angela and Abel, Jacob and Moung-Wen, Christine and others},
  journal={arXiv preprint arXiv:2405.12971},
  year={2024}
}

@inproceedings{liu2024grounding,
  title={Grounding dino: Marrying dino with grounded pre-training for open-set object detection},
  author={Liu, Shilong and Zeng, Zhaoyang and Ren, Tianhe and Li, Feng and Zhang, Hao and Yang, Jie and Jiang, Qing and Li, Chunyuan and Yang, Jianwei and Su, Hang and others},
  booktitle={European Conference on Computer Vision},
  pages={38--55},
  year={2024}
}

@inproceedings{wang2024llm,
  title={Llm-seg: Bridging image segmentation and large language model reasoning},
  author={Wang, Junchi and Ke, Lei},
  booktitle={Proceedings of the IEEE/CVF Conference on Computer Vision and Pattern Recognition},
  pages={1765--1774},
  year={2024}
}

@article{wang2024segllm,
  title={SegLLM: Multi-round Reasoning Segmentation},
  author={Wang, XuDong and Zhang, Shaolun and Li, Shufan and Kallidromitis, Konstantinos and Li, Kehan and Kato, Yusuke and Kozuka, Kazuki and Darrell, Trevor},
  journal={arXiv preprint arXiv:2410.18923},
  year={2024}
}

@inproceedings{jha2020doubleu,
  title={Doubleu-net: A deep convolutional neural network for medical image segmentation},
  author={Jha, Debesh and Riegler, Michael A and Johansen, Dag and Halvorsen, P{\aa}l and Johansen, H{\aa}vard D},
  booktitle={2020 IEEE 33rd International symposium on computer-based medical systems (CBMS)},
  pages={558--564},
  year={2020}
}

@article{ma2024segment,
  title={Segment anything in medical images},
  author={Ma, Jun and He, Yuting and Li, Feifei and Han, Lin and You, Chenyu and Wang, Bo},
  journal={Nature Communications},
  volume={15},
  number={1},
  pages={654},
  year={2024}
}

@misc{liu2024llavanext,
    title={LLaVA-NeXT: Improved reasoning, OCR, and world knowledge},
    url={https://llava-vl.github.io/blog/2024-01-30-llava-next/},
    author={Liu, Haotian and Li, Chunyuan and Li, Yuheng and Li, Bo and Zhang, Yuanhan and Shen, Sheng and Lee, Yong Jae},
    month={January},
    year={2024}
}

@misc{liu2023improvedllava,
      title={Improved Baselines with Visual Instruction Tuning}, 
      author={Liu, Haotian and Li, Chunyuan and Li, Yuheng and Lee, Yong Jae},
      publisher={arXiv:2310.03744},
      year={2023},
}

@article{GSCo,
  title={GSCo: Towards Generalizable AI in Medicine via Generalist-Specialist Collaboration},
  author={He, Sunan and Nie, Yuxiang and Wang, Hongmei and Yang, Shu and Wang, Yihui and Cai, Zhiyuan and Chen, Zhixuan and Xu, Yingxue and Luo, Luyang and Xiang, Huiling and others},
  journal={arXiv preprint arXiv:2404.15127},
  year={2024}
}

@article{MedSAM,
  title={Segment Anything in Medical Images},
  author={Ma, Jun and He, Yuting and Li, Feifei and Han, Lin and You, Chenyu and Wang, Bo},
  journal={Nature Communications},
  volume={15},
  pages={654},
  year={2024}
}

@article{chen2024huatuogpt,
  title={Huatuogpt-vision, towards injecting medical visual knowledge into multimodal llms at scale},
  author={Chen, Junying and Gui, Chi and Ouyang, Ruyi and Gao, Anningzhe and Chen, Shunian and Chen, Guiming Hardy and Wang, Xidong and Zhang, Ruifei and Cai, Zhenyang and Ji, Ke and others},
  journal={arXiv preprint arXiv:2406.19280},
  year={2024}
}

@article{tomar2022transresu,
  title={TransResU-Net: Transformer based ResU-Net for real-time colonoscopy polyp segmentation},
  author={Tomar, Nikhil Kumar and Shergill, Annie and Rieders, Brandon and Bagci, Ulas and Jha, Debesh},
  journal={arXiv preprint arXiv:2206.08985},
  year={2022}
}

@inproceedings{jha2019resunet++,
  title={Resunet++: An advanced architecture for medical image segmentation},
  author={Jha, Debesh and Smedsrud, Pia H and Riegler, Michael A and Johansen, Dag and De Lange, Thomas and Halvorsen, P{\aa}l and Johansen, H{\aa}vard D},
  booktitle={Proceedings of the 2019 IEEE International Symposium on Multimedia (ISM)},
  pages={225--2255},
  year={2019}
}

@article{isensee2018nnu,
  title={nnu-net: Self-adapting framework for u-net-based medical image segmentation},
  author={Isensee, Fabian and Petersen, Jens and Klein, Andre and Zimmerer, David and Jaeger, Paul F and Kohl, Simon and Wasserthal, Jakob and Koehler, Gregor and Norajitra, Tobias and Wirkert, Sebastian and others},
  journal={arXiv preprint arXiv:1809.10486},
  year={2018}
}

@inproceedings{cao2022swin,
  title={Swin-unet: Unet-like pure transformer for medical image segmentation},
  author={Cao, Hu and Wang, Yueyue and Chen, Joy and Jiang, Dongsheng and Zhang, Xiaopeng and Tian, Qi and Wang, Manning},
  booktitle={European conference on computer vision},
  pages={205--218},
  year={2022}
}

@misc{cheng2023sammed2d,
      title={SAM-Med2D}, 
      author={Junlong Cheng and Jin Ye and Zhongying Deng and Jianpin Chen and Tianbin Li and Haoyu Wang and Yanzhou Su and Ziyan Huang and Jilong Chen and Lei Jiangand Hui Sun and Junjun He and Shaoting Zhang and Min Zhu and Yu Qiao},
      year={2023},
      eprint={2308.16184},
      archivePrefix={arXiv},
      primaryClass={cs.CV}
}

@inproceedings{ronneberger2015u,
  title={U-net: Convolutional networks for biomedical image segmentation},
  author={Ronneberger, Olaf and Fischer, Philipp and Brox, Thomas},
  booktitle={Medical image computing and computer-assisted intervention--MICCAI 2015: 18th international conference, Munich, Germany, October 5-9, 2015, proceedings, part III 18},
  pages={234--241},
  year={2015}
}

@article{qwen2vl,
  title={Qwen2-vl: Enhancing vision-language model's perception of the world at any resolution},
  author={Wang, Peng and Bai, Shuai and Tan, Sinan and Wang, Shijie and Fan, Zhihao and Bai, Jinze and Chen, Keqin and Liu, Xuejing and Wang, Jialin and Ge, Wenbin and others},
  journal={arXiv preprint arXiv:2409.12191},
  year={2024}
}

@article{llama,
  title={Llama: Open and efficient foundation language models},
  author={Touvron, Hugo and Lavril, Thibaut and Izacard, Gautier and Martinet, Xavier and Lachaux, Marie-Anne and Lacroix, Timoth{\'e}e and Rozi{\`e}re, Baptiste and Goyal, Naman and Hambro, Eric and Azhar, Faisal and others},
  journal={arXiv preprint arXiv:2302.13971},
  year={2023}
}

@inproceedings{kirillov2023segment,
  title={Segment anything},
  author={Kirillov, Alexander and Mintun, Eric and Ravi, Nikhila and Mao, Hanzi and Rolland, Chloe and Gustafson, Laura and Xiao, Tete and Whitehead, Spencer and Berg, Alexander C and Lo, Wan-Yen and others},
  booktitle={Proceedings of the IEEE/CVF International Conference on Computer Vision},
  pages={4015--4026},
  year={2023}
}
}


\end{document}